\documentclass[10pt,twocolumn,letterpaper]{article}

\usepackage{amsmath,amsfonts,bm}

\def\eqref#1{equation~\ref{#1}}

\def\1{\bm{1}}

\def\mD{{\bm{D}}}
\def\mE{{\bm{E}}}
\def\mF{{\bm{F}}}

\def\mK{{\bm{K}}}

\def\mQ{{\bm{Q}}}

\def\mV{{\bm{V}}}

\DeclareMathAlphabet{\mathsfit}{\encodingdefault}{\sfdefault}{m}{sl}
\SetMathAlphabet{\mathsfit}{bold}{\encodingdefault}{\sfdefault}{bx}{n}

\newcommand{\R}{\mathbb{R}}

\usepackage{iccv}
\usepackage{times}
\usepackage{epsfig}
\usepackage{graphicx}
\usepackage{amsmath}
\usepackage{amssymb}
\usepackage{bbding}
\usepackage{makecell}
\usepackage{subcaption}
\usepackage{caption}
\usepackage{color}
\definecolor{gray}{RGB}{200, 200, 200}
\newcommand{\rulesep}{\unskip\hfill{\color{gray}\vrule}\hfill\ignorespaces}

\usepackage[breaklinks=true,bookmarks=false]{hyperref}

\iccvfinalcopy 

\begin{document}

\title{Long-Range Grouping Transformer for Multi-View 3D Reconstruction}

\author{Liying Yang$^{*}$\hspace{0.05in}
Zhenwei Zhu$^{*}$\hspace{0.05in}
Xuxin Lin\hspace{0.05in}
Jian Nong\hspace{0.05in}
Yanyan Liang$^{\dag}$\\
Macau University of Science and Technology\hspace{0.1in} \\
}
\maketitle

\renewcommand{\thefootnote}{\fnsymbol{footnote}}
\footnotetext[1]{Equal contribution. Email: \{lyyang69, garyzhu1996\}@gmail.com}
\footnotetext[2]{Corresponding author. Email: yyliang@must.edu.mo}

\begin{abstract}
Nowadays, transformer networks have demonstrated superior performance in many computer vision tasks. In a multi-view 3D reconstruction algorithm following this paradigm, self-attention processing has to deal with intricate image tokens including massive information when facing heavy amounts of view input. The curse of information content leads to the extreme difficulty of model learning. To alleviate this problem, recent methods compress the token number representing each view or discard the attention operations between the tokens from different views. Obviously, they give a negative impact on performance. Therefore, we propose long-range grouping attention (LGA) based on the divide-and-conquer principle. Tokens from all views are grouped for separate attention operations. The tokens in each group are sampled from all views and can provide macro representation for the resided view. The richness of feature learning is guaranteed by the diversity among different groups. An effective and efficient encoder can be established which connects inter-view features using LGA and extract intra-view features using the standard self-attention layer. Moreover, a novel progressive upsampling decoder is also designed for voxel generation with relatively high resolution. Hinging on the above, we construct a powerful transformer-based network, called LRGT. Experimental results on ShapeNet verify our method achieves SOTA accuracy in multi-view reconstruction. Code will be available at \url{https://github.com/LiyingCV/Long-Range-Grouping-Transformer}.
\end{abstract}

\section{Introduction}
\label{Introduction}
3D reconstruction, the problem to recover the shape of an object according to its several view images, is an essential research topic involving the field of computer vision and computer graphics. One of the main challenges is how to extract the features from multiple images for generating the corresponding object with 3D representation. At present, deep learning reconstructors have provided three kinds of solutions, including RNN-based methods \cite{choy20163d, kar2017learning}, CNN-based methods \cite{su2015multi, wang20173densinet, huang2018deepmvs, paschalidou2018raynet, xie2019pix2vox, xie2020pix2vox++, yang2020robust, zhu2023garnet} and transformer-based methods \cite{wang2021multi, yagubbayli2021legoformer, shi20213d, zhu2023umi}. In this work, we focus on the transformer network and aim to improve the accuracy and robustness of multi-view 3D reconstruction with voxel representation.

\begin{figure*}
  \centering
  \begin{subfigure}{0.226\linewidth}
    \scalebox{1.25}{
    \includegraphics[width=0.7\linewidth]{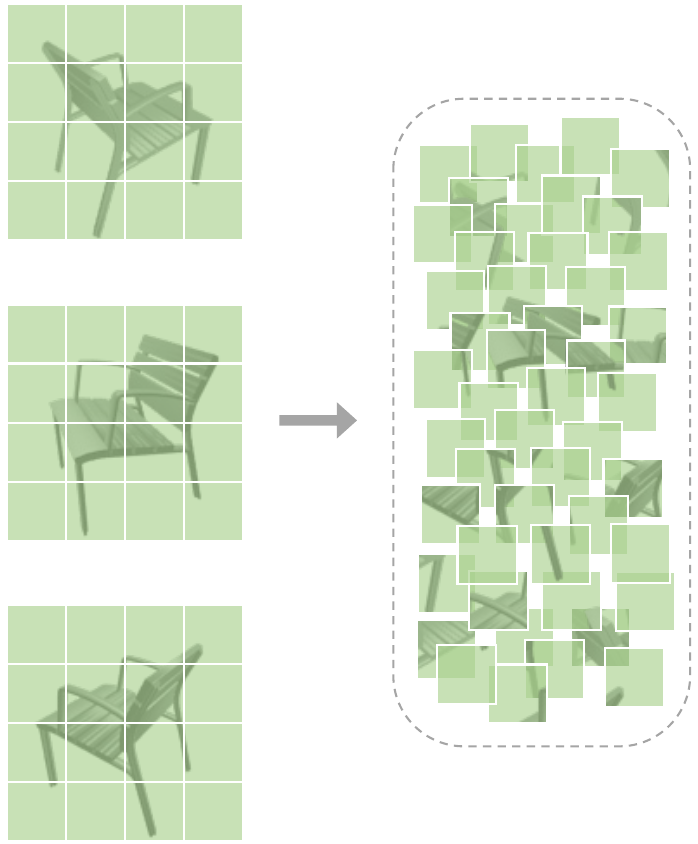}}
    \caption{FRA}
    \label{full_range}
  \end{subfigure}
  \rulesep
  \begin{subfigure}{0.222\linewidth}
   \scalebox{1.25}{
    \includegraphics[width=0.7\linewidth]{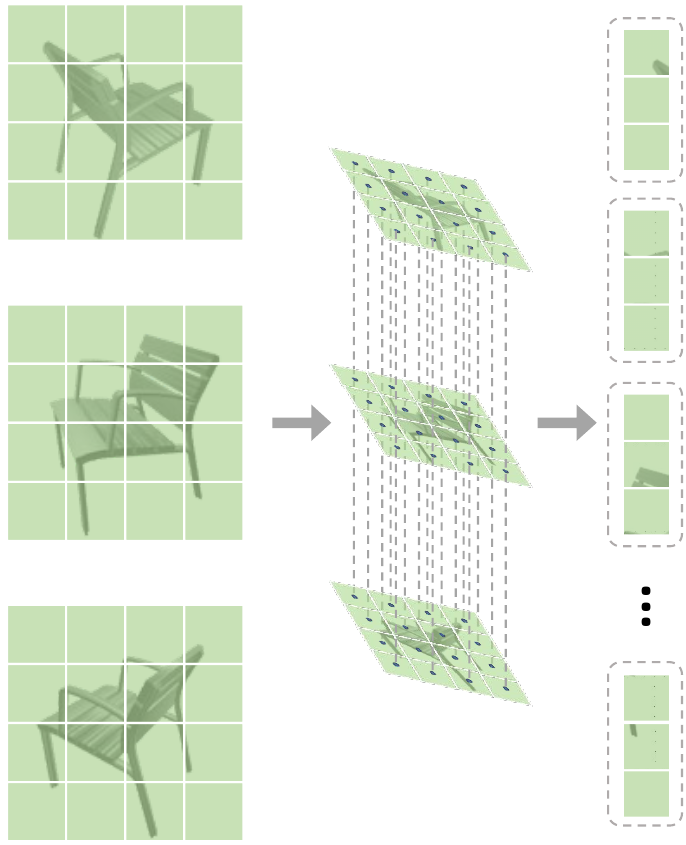}}
    \caption{TGA}
    \label{token_range}
  \end{subfigure}
  \rulesep
  \begin{subfigure}{0.26\linewidth}
    \scalebox{1.25}{
    \includegraphics[width=0.725\linewidth]{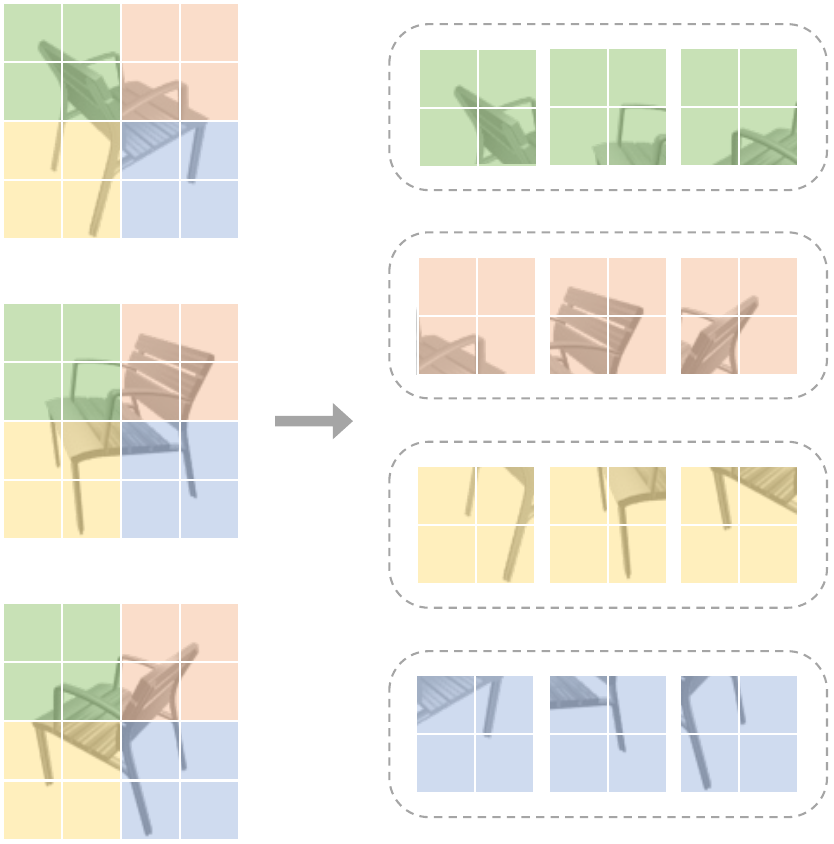}}
    \caption{SGA}
    \label{short_range}
  \end{subfigure}
  \rulesep
  \begin{subfigure}{0.26\linewidth}
   \scalebox{1.25}{
    \includegraphics[width=0.725\linewidth]{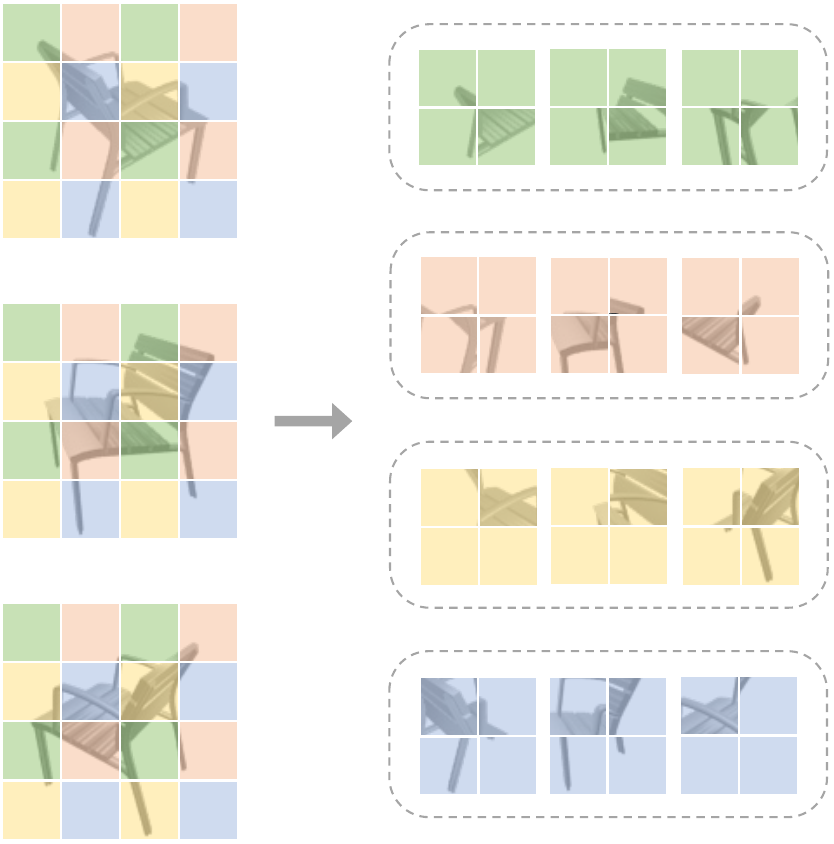}}
    \caption{LGA (Ours)}
    \label{long_range}
  \end{subfigure}
  \caption{Illustration of different attention strategies for processing multi-view input. (a) Full-Range Attention (FRA); (b) Token-Range Grouping Attention (TGA); (c) Short-Range Grouping Attention (SGA); (d) Long-Range Grouping Attention (LGA). Tokens in the same dashed box or on the same dashed line are divided into the same group for attention operation. It means that they will build the correlations.}
  \label{highlight}
\end{figure*}

Vision transformer (ViT) \cite{dosovitskiy2021image} promotes the approach to extract features from an image employing transformer architecture. An image is split into fixed-size patches and then using attention operators learn the association between them to explore an appropriate representation. If following this paradigm to establish a multi-view 3D reconstruction algorithm, an intuitive approach like \cite{zeng2020learning} is to build attention layers to connect full-range tokens from all views, shown in Figure~\ref{full_range}. However, it is extremely difficult for the model to face heavy amounts of view input, as attention operators have to predict the potential importance weights of intricate tokens from views. The curse of information content increases the complexity of model encoding. It is necessary to provide relatively large-scale data to support adequate training, but the commonly used datasets for multi-view reconstruction are far from satisfying such requirements.

The previous works design three types of strategies to avoid this problem. 1) Separated-stage strategy \cite{shi20213d} individually extract the feature from each view and then fusion them. It is simplified in structure but weak in mining the inter-view relationships. 2) Blended-stage strategy \cite{wang2021multi, yagubbayli2021legoformer} compress the number of tokens to represent each view and make the attention layers to learn the association between all tokens from different views. They enhance the inter-view correlation but lose a certain representation capability for each view. 3) Alternated-stage strategy \cite{zhu2023umi} employs decoupling to alternate the feature extraction from intra-view and inter-view tokens. They discard the attention operations between the tokens from different views but insert other modules to achieve similar purposes, which brings extra computation. The efficiency is significantly dragged down, especially for a large number of image inputs.

Considering these shortcomings, we hold that grouping tokens in attention operations is a more reasonable solution. The role of full-range attention can be equivalent to a combination of two attention layers. ViT can be used as the basic architecture of encoder independent and parallel processing on each view, while some of the original attention layers are replaced by special grouping attention operations for multi-branch. Adopting the principle of divide-and-conquer, the tokens from different views are divided into some groups under certain rules. The amount of tokens processed by the attention layer is greatly reduced, thus alleviating the learning difficulty of the model.

For the ambiguous associations between view inputs in the multi-view reconstruction task, it is necessary to select a suitable grouping strategy for the special attention operation. 
There are some different grouping strategies in the previous research works about transformer. The token-range grouping attention as shown in Figure~\ref{token_range}, which assigns tokens with the same location from different views as a group, ignores the auxiliary role of the intra-view token-to-token association in building inter-view dependencies. The short-range grouping attention as shown in Figure~\ref{short_range} (e.g., \cite{liu2021decoupled}) increases the connectivity between local tokens in each view. However, it is hard to construct long-distance associations and only reinforces the sensitivity of tokens towards local information, hence it is applicable to multi-input problems with temporal coherence such as video.

For our task, we propose the long-range grouping attention (LGA) as shown in Figure~\ref{long_range}. The tokens from the same view in each group can provide a certain macro representation for their view but are not limited to local features. To further enhance differences between the tokens from different views, we also introduce the inter-view feature signatures (IFS) to assist the attention processing. In addition, we propose a novel progressive upsampling decoder that integrates the transposed convolution into the transformer block. It fully exploits the self-attention to mine features in the relatively high-resolution voxel space.

The contributions can be summarized as follows:
\begin{itemize}
\item{We propose the long-range grouping attention which can simply and efficiently establish the correlations between different images. A novel encoder for handling multi-view input is formed by integrating this attention operator into a standard transformer architecture.}
\item{We overcome the difficulty of transformer block working on the relatively high-resolution voxel directly to propose a progressive upsampling decoder that can strengthen the reconstruction ability.}
\item{Experimental results on ShapeNet \cite{chang2015shapenet} demonstrate that our method outperforms other SOTA methods in multi-view reconstruction. Additional experiment results on Pix3D~\cite{sun2018pix3d} also verify its effectiveness on real-world data.}
\end{itemize}

\begin{figure*}
  \centering
  \begin{subfigure}{.95\linewidth}
    \scalebox{1.}{
    \includegraphics[width=1.\linewidth]{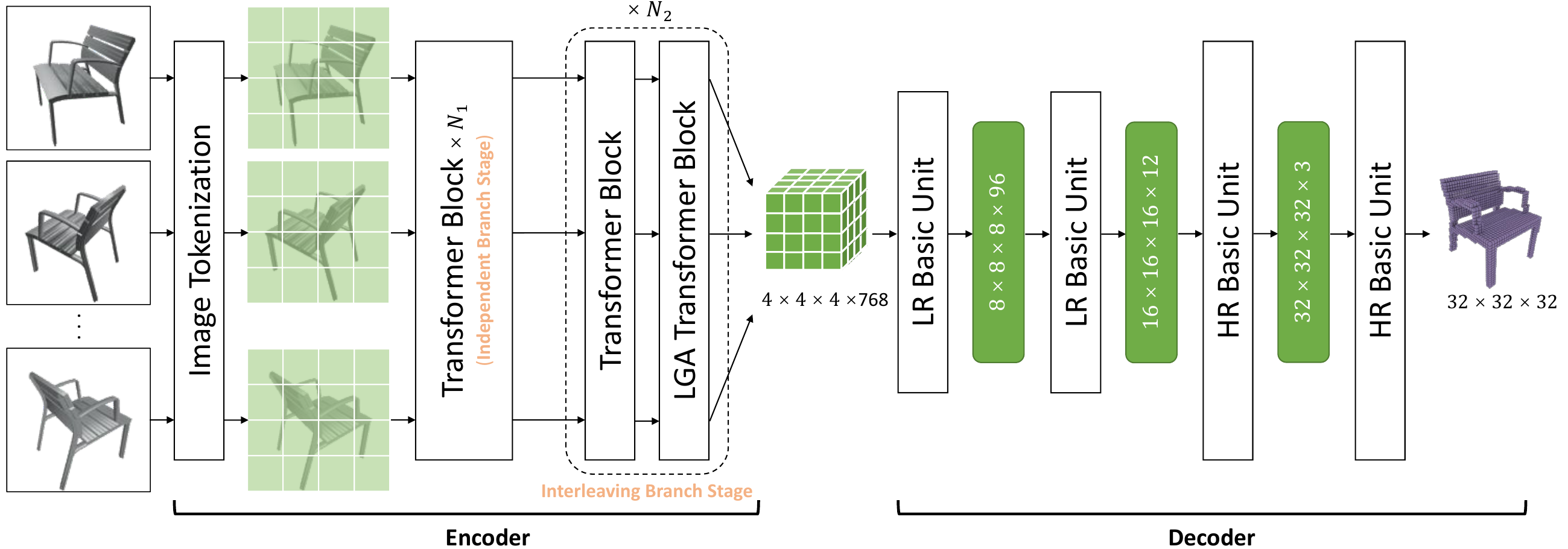}}
    \caption{Overall Network Architecture}
    \label{overall}
  \end{subfigure} \vskip 1mm
  \begin{subfigure}{0.345\linewidth}
   \scalebox{1.}{
    \includegraphics[width=1.\linewidth]{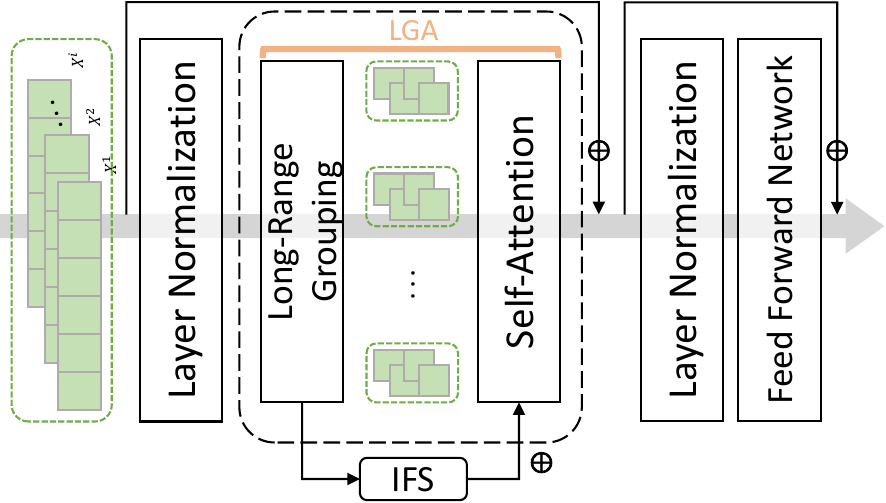}}
    \caption{LGA Transformer Block}
    \label{lga_trans_block}
  \end{subfigure}
  \rulesep
  \begin{subfigure}{0.19\linewidth}
    \scalebox{1.}{
    \includegraphics[width=1.\linewidth]{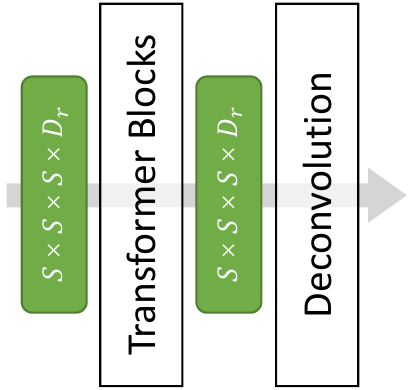}}
    \caption{LR Basic Unit}
    \label{lr_basic_unit}
  \end{subfigure}
  \rulesep
  \begin{subfigure}{0.35\linewidth}
   \scalebox{1.}{
    \includegraphics[width=1.\linewidth]{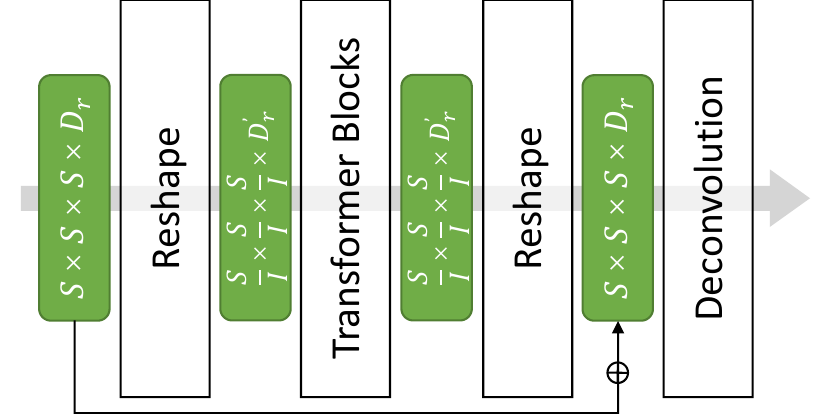}}
    \caption{HR Basic Unit}
    \label{hr_basic_unit}
  \end{subfigure}
  \caption{Illustration of our proposed LRGT and its details. Among them, the IFS module only includes a convolutional layer.}
  \label{highlight}
\end{figure*}

\section{Related Works}

\subsection{Multi-View 3D Reconstruction}
Traditional methods such as SfM~\cite{ozyecsil2017survey} and SLAM~\cite{fuentes2015visual} reconstruct scenes by matching features and estimating camera pose. However, they struggle to handle images in complex situations. Recently, deep learning-based methods have gained significant attention in multi-view 3D reconstruction.

In RNN-based and CNN-based methods, the encoder and fusion modules are separated. 3D-R2N2~\cite{choy20163d} and LSM~\cite{kar2017learning} use RNN to fuse multi-view features. However, RNNs are permutation-invariant and time-consuming. 3DensiNet~\cite{wang20173densinet} aggregates multi-view features using max-pooling, but this approach cannot establish a learnable view connection. Attsets~\cite{yang2020robust}, Pix2Vox~\cite{xie2019pix2vox,xie2020pix2vox++}, and GARNet~\cite{zhu2023garnet} employ attention layer to capture inter-view correlations. Although these methods have achieved promising results, they are weak for mining the inter-view relationships because the relevant structure is too simple.

In transformer-based methods, EVolT~\cite{wang2021multi} and LegoFormer~\cite{yagubbayli2021legoformer} compress the number of tokens to represent each view using CNN-backbone network, in order to integrate feature extraction and view feature fusion into a single transformer network. However, they lose a certain representation capability for each view. On the other hand, 3D-RETR~\cite{shi20213d} handles each view employing the transformer network independently and fuses the features using the pooling method. In fact, it is the same structure that separates encoder and fusion as aforementioned, which ignores multi-view associations. UMIFormer~\cite{zhu2023umi} adopts decoupled intra- and inter-view feature extraction to strengthen the correlation between views, whereas it introduces additional modules to estimate inter-view relationships.

Different from the above methods that focus on solving multi-view fusion problems, the optimization-based methods, such as BARF~\cite{lin2021barf}, NeRS~\cite{zhang2021ners}, and FvOR~\cite{yang2022fvor}, jointly optimize 3D shape and camera pose estimation. They have good performance but sacrifice efficiency.

\subsection{Group Self-Attention}
\cite{vaswani2017attention} proposes a transformer network and makes success in natural language processing. In order to migrate this method to process images, \cite{Fwang2020axial} divides the 2D attention into two groups based on the height and width dimensions. \cite{huang2019interlaced} employs both long-range and short-range methods to construct multiple feature groups from a single image. Recently, ViT~\cite{dosovitskiy2021image} which processes an image by splitting it into patches has demonstrated superior performance in many vision tasks. STTN~\cite{zeng2020learning} introduces ViT into video inpainting and serves as a pioneering attempt to apply the transformer architecture to address a problem with multi-view inputs. It utilizes full-range attention for all tokens in the spatial-temporal dimension. To mitigate the difficulty to process massive information, DSTT~\cite{liu2021decoupled} employs the short-range grouping attention. It allows the model to capture the local features while paying less attention to the non-local relationships. Therefore, this grouping strategy is suitable for addressing the problem with multi-input with temporal coherence such as video. However, for our task with ambiguous associations between views, it requires a method to establish long-distance dependency relationships.

\section{Methods}

The overall framework of our proposed method is illustrated in Figure~\ref{overall}. The view image set of an object $\mathcal{I}=\{I_1,I_2,\ldots,I_N\}$ is processed by the encoder $\mE$ to extract the feature representation for reconstruction. Then the decoder $\mD$ generates the corresponding voxel-based 3D shape $O$. The overall process is formulated as:
\begin{equation}
  O = \mathtt{LRGT}(\mathcal{I}) = \mD(\mE(\mathcal{I})). 
\end{equation}

\subsection{Encoder}

The encoder is based on the architecture of ViT~\cite{dosovitskiy2021image} and is divided into two stages: an independent branch stage and an interleaving branch stage. The first stage consists of $N_1$ standard transformer blocks. The second stage including $N_2$ basic unit alternately handles intra-view tokens and inter-view tokens separately. The former still employs standard transformer blocks to parallelize all views, while the latter utilizes transformer blocks with the long-range grouping attention (LGA) to establish communication between views and inter-view feature signatures (IFS) to help distinguish the feature of different views as shown in Figure~\ref{lga_trans_block}. At last, we introduce the similar-token merger~\cite{zhu2023umi} to compress the tokens from all branches to the specified size.

\textbf{Long-Range Grouping Attention.} This is a special attention layer for establishing the correlations between different views. Given specific $i$-th view feature $X^i\in \R^{P\times D_e}$, where $P$ is the number of tokens and $D_e$ is the token dimension. We perform uniform and regular sampling based on the length and width of $X^i$ to obtain $g$ groups. It is noteworthy that tokens in each group are not adjacent to each other in terms of the original view features. Rewrite $X^i$ as $X^i=\{x_1^i,\ldots,x_j^i\}$, $i=1,\ldots,N$, where $x_j^i$ represents the tokens from $j$-th group $i$-th view, and $N$ represents the amount of input view. As shown in Figure~\ref{long_range}, the tokens sharing the same color belong to a group, which implies that each group not only includes intra-view tokens but also contains the inter-view tokens from their corresponding positions. After grouping, self-attention operations process the tokens in each group independently. Essentially, this method utilizes the principle of divide-and-conquer to reduce the difficulty of model training. The attention operations for all groups are predication about the relationship of views, but there are certain differences between them. Feature diversity is beneficial for building high-quality representations.

\textbf{Inter-View Feature Signatures.} The multi-view reconstruction task is invariant to permutations, hence the positional encoding we set is the same for all views. It may cause the LGA to neglect the relationship whether some tokens are from the different view. Inspire by~\cite{dong2022cswin}, we introduce a simple structure that only includes a convolutional layer as IFS to enhance differences between tokens from various views. Given series tokens $x_j^i$ from $j$-th group $i$-th view, we obtain the token features by a trainable linear projection $\phi$. The $\phi$ shares the linear projection which is used to predict $\mV_j$ in this transformer block, where $\phi(x_j^i)\in \mathbb{R}^{\frac{P}{g}\times D_e}$, and $g$, $P$, $D_e$ represent the number of groups, the number of tokens and token dimension. The feature signatures $f_j^i$ from $j$ group $i$ view is obtained as follow:
\begin{equation}
  \label{equ:sig}
  f_j^i=\mathtt{IFS}(\phi(x_j^i)),
\end{equation}
Furthermore, we define inter-view feature signatures $\mF_j=\{f_j^1,\ldots,f_j^N\}$, $j=1,\ldots,g$, where $N$ represent the number of views. Formally, for each view features $x_j^i$ from $X^i$, the LGA with IFS becomes:
\begin{equation}
  \label{equ:att}
  \mathtt{Attn}(\mQ_j,\mK_j,\mV_j) = \mathtt{Softmax}(\frac{\mQ_j\mK_j^T}{\sqrt d})\mV_j+\mF_j,
\end{equation}
Where $\mQ_j,\mK_j,\mV_j$ represent queries, keys, values from $X_j=\{x_j^1, \dots ,x_j^N\}$ respectively, and $d$ is the channel number of the queries. The attention maps obtained from individual groups in the group dimension are concatenated, thereby completing an inter-view association computation.

\subsection{Decoder}
In our investigation, the previous transformer-based decoders for voxel reconstruction employ transformer blocks to generate a low-resolution (LR) feature representation (e.g., $4^3$) and then upsample to the target size in hasty steps. Because the relatively high-resolution (HR) tensor corresponds to a large number of tokens which is unbearable for attention operations. However, it is difficult to reconstruct details through a rapid upsampling approach on the last few layers of the model. To overcome the limitations, we design a progressive upsampling transformer-based decoder.

Our proposed decoder upsamples the feature map of size $4^3$ extracted by the encoder to $32^3$ gradually through 4 basic units. Each basic unit includes 1 deconvolution layer and 2 transformer blocks. Convolution and multi-head self-attention and processing are complementary~\cite{park2022how}. Deconvolution is better at encoding local information and supports upsampling with relatively low consumption, while the transformer has excellent global representation capabilities. Therefore, we hold that combining the advantages of both can build a powerful model.

The structure of a basic unit can easily handle LR feature inputs in the way shown in Figure~\ref{lr_basic_unit}. However, as aforementioned, the transformer block cannot normally deal with a large number of tokens brought by HR features. To this end, we modify it to be an HR basic unit as shown in Figure~\ref{hr_basic_unit}.

In the HR basic unit, the adjacent voxels are uniformly grouped and sequentially combined to compress the number of tokens. Given voxel features $V=\{v_1,\ldots, v_t \}$, $V\in \R^{T\times D_r}$, where $v_t$ is $t$-th voxel token, $T, D_r$ represent the number of token numbers and token dimension, respectively. In the resolution of $S\times S\times S$, we define $T= S\times\ S \times S $. Grouping every $I\times I \times I$ adjacent voxel token together, we obtain the voxel features $V\in\R^{T’ \times {D_r}'}$, where $T’=\frac{S}{I} \times \frac{S}{I} \times \frac{S}{I}$, ${D_r}'=I^3 \times D_r$. As a result, voxel features $V$ are grouped into sizes that can be processed by self-attention while preserving the original feature information in a relatively high resolution (e.g., $16^3$ and $32^3$). In the end, we take voxel features $V$ as input and apply multi-head self-attention to explore voxel location relationships.

It is worth noting that the attention layer under such a method loses certain concern about the relevance of voxels assigned to the same token. We consider that the convolution layers in the network which focus on the local feature can be used to supplement the lack, so we add a skip connection to combine the features represented by the deconvolution and transformer layer.

\subsection{Loss function}
Following~\cite{shi20213d}, we use Dice loss \cite{milletari2016v} as the loss function which it is suitable for high unbalanced voxel occupancy. The Dice loss could be formulated as:
\begin{equation}
    \mathcal{L} = 1-\frac{\sum_{i=1}^{32^3}p_{i} g_{i}}{\sum_{i=1}^{32^3}p_{i}+g_{i}}-\frac{\sum_{i=1}^{32^3}\left(1-p_{i}\right)\left(1-g_{i}\right)}{\sum_{i=1}^{32^3}2-p_{i}-g_{i}}.
\end{equation}
where $p_i$ and $g_i$ represent the confidence of $i$-th voxel grid on the reconstructed volume and ground truth.

\begin{table*}[]
	\centering
	\renewcommand\arraystretch{1.6}
	\scalebox{0.63}{
	\begin{tabular}{c|ccccccccc}
	    \hline
		\multicolumn{1}{c|}{\textbf{Methods}} & \multicolumn{1}{c}{\textbf{1 view}} & \multicolumn{1}{c}{\textbf{2 views}} & \multicolumn{1}{c}{\textbf{3 views}} & \multicolumn{1}{c}{\textbf{4 views}} & \multicolumn{1}{c}{\textbf{5 views}} & \multicolumn{1}{c}{\textbf{8 views}} & \multicolumn{1}{c}{\textbf{12 views}} & \multicolumn{1}{c}{\textbf{16 views}} & \multicolumn{1}{c}{\textbf{20 views}} \\ \hline
		\textbf{3D-R2N2 \cite{choy20163d}} & 0.560 / 0.351 & 0.603 / 0.368 & 0.617 / 0.372 & 0.625 / 0.378 & 0.634 / 0.382 & 0.635 / 0.383 & 0.636 / 0.382 & 0.636 / 0.382 & 0.636 / 0.383\\
		\textbf{AttSets \cite{yang2020robust}} & 0.642 / 0.395 & 0.662 / 0.418 & 0.670 / 0.426 & 0.675 / 0.430 & 0.677 / 0.432 & 0.685 / 0.444 & 0.688 / 0.445 & 0.692 / 0.447 & 0.693 / 0.448 \\
		\textbf{Pix2Vox++ \cite{xie2020pix2vox++}} & 0.670 / 0.436 & 0.695 / 0.452 & 0.704 / 0.455 & 0.708 / 0.457 & 0.711 / 0.458 & 0.715 / 0.459 & 0.717 / 0.460 & 0.718 / 0.461 & 0.719 / 0.462 \\
        \textbf{GARNet \cite{zhu2023garnet}} & 0.673 / 0.418 & 0.705 / 0.455 & 0.716 / 0.468 & 0.722 / 0.475 & 0.726 / 0.479 & 0.731 / 0.486 & 0.734 / 0.489 & 0.736 / 0.491 & 0.737 / 0.492 \\
		\textbf{GARNet+} & 0.655 / 0.399 & 0.696 / 0.446 & 0.712 / 0.465 & 0.719 / 0.475 & 0.725 / 0.481 & 0.733 / 0.491 & 0.737 / 0.498 & 0.740 / 0.501 & 0.742 / 0.504 \\ \hline
		\textbf{EVolT \cite{wang2021multi}} & - / - & - / - & - / - & 0.609 / 0.358 & - / - & 0.698 / 0.448 & 0.720 / 0.475 & 0.729 / 0.486 & 0.735 / 0.492 \\
		\textbf{Legoformer \cite{yagubbayli2021legoformer}} & 0.519 / 0.282 & 0.644 / 0.392 & 0.679 / 0.428 & 0.694 / 0.444 & 0.703 / 0.453 & 0.713 / 0.464 & 0.717 / 0.470 & 0.719 / 0.472 & 0.721 / 0.472 \\ 
        \textbf{3D-RETR \cite{shi20213d}} (3 view) & 0.674 / - & 0.707 / - & 0.716 / - & 0.720 / - & 0.723 / - & 0.727 / - & 0.729 / - & 0.730 / - & 0.731 / - \\
        \textbf{UMIFormer \cite{zhu2023umi}} & 0.6802 / 0.4281 & 0.7384 / 0.4919 & 0.7518 / 0.5067 & 0.7573 / 0.5127 & 0.7612 / 0.5168 & 0.7661 / 0.5213 & 0.7682 / 0.5232 & 0.7696 / 0.5245 & 0.7702 / 0.5251 \\
        \textbf{UMIFormer+} & 0.5672 / 0.3177 & 0.7115 / 0.4568 & 0.7447 / 0.4947 & 0.7588 / 0.5104 & 0.7681 / 0.5216 & 0.7790 / 0.5348 & 0.7843 / 0.5415 & 0.7873 / 0.5451 & 0.7886 / 0.5466 \\ \hline 
        \textbf{LRGT (Ours)} & \textbf{0.6962} / \textbf{0.4461} & \textbf{0.7462} / \textbf{0.5005} & \textbf{0.7590} / \textbf{0.5148} & \textbf{0.7653} / \textbf{0.5214} & 0.7692 / 0.5257 & 0.7744 / 0.5311 & 0.7766 / 0.5337 & 0.7781 / 0.5347 & 0.7786 / 0.5353 \\
        \textbf{LRGT+ (Ours)} & 0.5847 / 0.3378 & 0.7145 / 0.4618 & 0.7476 / 0.4989 & 0.7625 / 0.5161 & \textbf{0.7719} / \textbf{0.5271} & \textbf{0.7833} / \textbf{0.5403} & \textbf{0.7888} / \textbf{0.5467} & \textbf{0.7912} / \textbf{0.5497} & \textbf{0.7922} / \textbf{0.5510} \\ \hline
	\end{tabular}}
	\caption{Evaluations of multi-view 3D reconstruction results on ShapeNet using IoU / F-score$@1\%$. The best score is highlighted in bold.}
\label{total_result}
\end{table*}

\section{Experiments}
\subsection{Datasets}
The experiments are performed on ShapeNet dataset~\cite{chang2015shapenet}. Following~\cite{choy20163d}, we utilize a subset of ShapeNet consisting of 13 categories and 43,783 3D objects for a fair comparison. Additionally, we evaluate our model on real-world data from the Pix3D dataset~\cite{sun2018pix3d}, which includes 2,894 untruncated and unoccluded chair images following~\cite{xie2019pix2vox,xie2020pix2vox++}.

\subsection{Metrics} Following the same evaluation methods as the current advance works, we measure the reconstruction quality of our models using the mean Intersection-over-Union (IoU) and F-score$@1\%$~\cite{xie2020pix2vox++}. Higher IoU and F-score$@1\%$ values indicate better performance.

\textbf{IoU:} The mean Intersection-over-Union (IoU) is formulated as:
\begin{equation}
	\text{IoU}=\frac{\sum_{(i,j,k)}\text{I}(\hat{p}(i,j,k)>t)\text{I}(p(i,j,k))}{\sum_{(i,j,k)}\text{I}[\text{I}(\hat{p}(i,j,k)>t)+\text{I}(p(i,j,k))]}, \label{iou}
\end{equation}
where $\hat{p}(i,j,k)$ and $p(i,j,k)$ represent the predicted occupancy probability and ground truth at $(i,j,k)$. $\text{I}(\cdot)$ is an indicator function and $t$ denotes a voxelization threshold.

\textbf{F-Score$@1\%$:} Following the same setting as~\cite{xie2020pix2vox++}, we take F-Score~\cite{tatarchenko2019single} as an extra metric to evaluate the performance of 3D reconstruction results, which can be formulated as 
\begin{equation}
	\text{F-Score}(d)=\frac{2\text{P}(d)\text{R}(d)}{\text{P}(d)+\text{R}(d)}, \label{fscore}
\end{equation}
where $\text{P}(d)$ and $\text{R}(d)$ denote the precision and recall for a distance threshold between prediction and ground truth. More formally, 
 
\begin{equation}
    \text{P}(d)=\frac{1}{n_{\mathcal{R}}}\sum_{r\in \mathcal{R}}[{min}_{g\in \mathcal{G}}||\text{r}-\text{g}||<d],
\end{equation}
\begin{equation}
    \text{R}(d)=\frac{1}{n_{\mathcal{R}}}\sum_{r\in \mathcal{G}}[{min}_{r\in \mathcal{R}}||\text{g}-\text{r}||<d],
\end{equation}
where $[\cdot]$ is the iverson bracket. $\mathcal{R}$ and $\mathcal{G}$ denote the reconstructed and ground truth point clouds, respectively. $n_{\mathcal{R}}$ and $n_{\mathcal{G}}$ are the numbers of points in $\mathcal{R}$ and $\mathcal{G}$. For voxel representation, we generate the object surface by the marching cubes algorithm \cite{lorensen1987marching}. Then 8,192 points are sampled from the object surface to compute F-Score between prediction and ground truth. F-Score$@1\%$ indicates the F-Score value when $d$ is set to $1\%$.

\begin{figure*}[t]
    \centering
	\begin{minipage}{1\linewidth}
        \centerline{\includegraphics[width=1\linewidth]{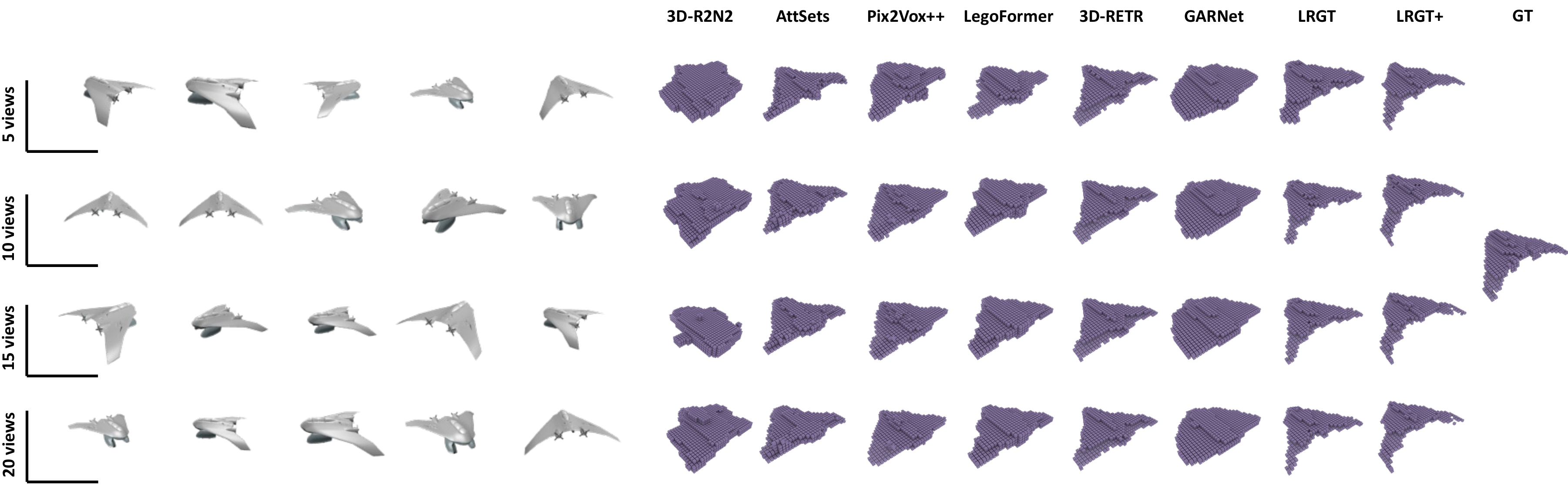}}
	\end{minipage}
	\\ \vspace{0.5mm}
	\begin{minipage}{1\linewidth}
        \centerline{\includegraphics[width=1\linewidth]{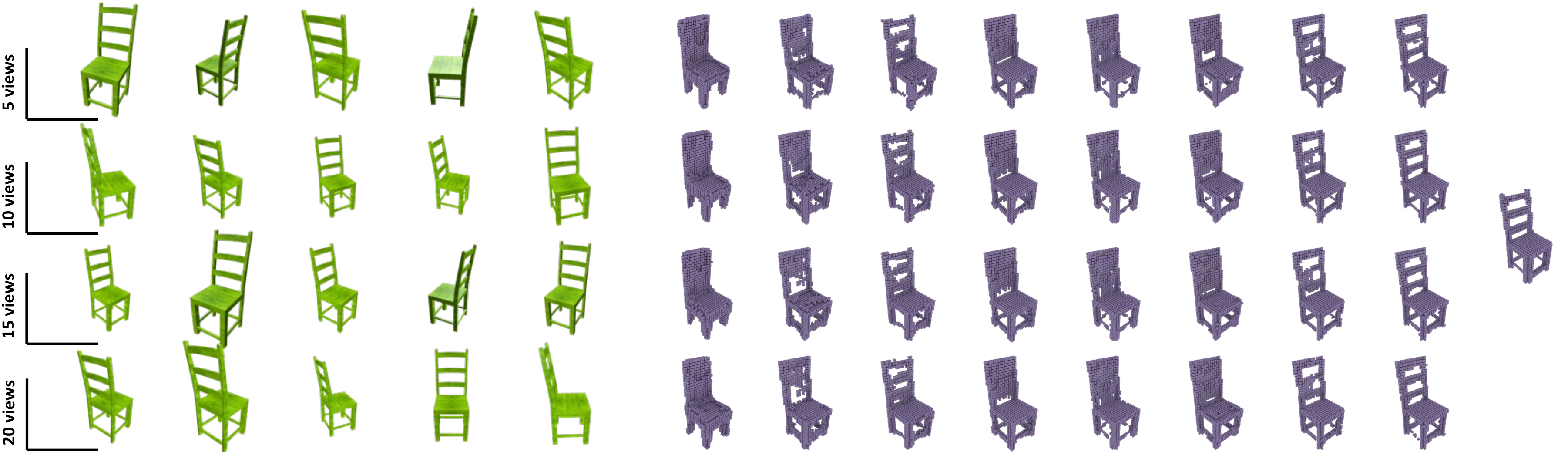}}
	\end{minipage}
	\caption{Multi-view reconstruction results on the test set of ShapeNet when facing 5 views, 10 views, 15 views and 20 views as input.}
\label{show_results}
\end{figure*}

\subsection{Implementation Details}

We utilize the pre-training model of DeiT-B~\cite{touvron2021training} to initialize our encoder and the cls token and distillation token are removed. Our encoder consists of 12 transformer blocks where $N_{1}=6$ and $N_{2}=3$. In LGA Transformer blocks, the $g$ is defined as 49 in LGA. For our decoder, the $I$ is 2 and 4 in the first and second HR basic unit, respectively. To validate the reconstruction performance of the model, we provide two models with the same structure namely LRGT and LRGT+ which use 3 and 8 views as input during training following~\cite{zhu2023garnet, zhu2023umi}. During inference, the models can adapt to an arbitrary number of image inputs. In detail, following~\cite{xie2020pix2vox++}, we use $224\times 224$ RGB images as input and set the voxelized output to $32 \times 32 \times 32$. LRGT and LRGT+ are trained by an AdamW optimizer~\cite{loshchilov2018decoupled} with a $\beta_1$ of 0.9 and a $\beta_2$ of 0.999. The training batch size is 32 for 110 epochs. The learning rate is set to 0.0001 and sequentially decayed by 0.1 after 60 and 90 epochs. We use a threshold of 0.5 for LRGT and 0.4 for LRGT+ to obtain the occupancy voxel grid. Notably, the experiments in Section~\ref{experiments::strategy} and \ref{experiments::ablation} take 3-view-input during training as an instance.

\subsection{Multi-view 3D Object Reconstruction}
\textbf{Quantitative results.} We compare LRGT and LRGT+ with the SOTA methods~\cite{choy20163d,yang2020robust,xie2020pix2vox++,wang2021multi,yagubbayli2021legoformer,shi20213d,zhu2023umi,zhu2023garnet} on the ShapeNet dataset and the results are shown in Table~\ref{total_result}. Obviously, LRGT significantly outperforms the previous methods in all evaluation metrics for both single-view and multi-view reconstruction. LRGT+ degrades part of the performance for single-view reconstruction while further improving the ability to process multi-view inputs.

\textbf{Qualitative results.} Figure~\ref{show_results} shows two sets of reconstruction examples using different methods according to varying numbers of input views on the test set of the ShapeNet dataset. LRGT and LRGT+ outperform the other methods in restoring the overall shape and fine-grained details of the object. Compared with other methods, our model captures more accurate contours of the airplane and restores the chair details better especially the backrest.

\subsection{Grouping Attention Strategy}
\label{experiments::strategy}
\begin{figure}[t]
	\centering
	\scalebox{1.25}{
	\includegraphics[width=0.8\linewidth]{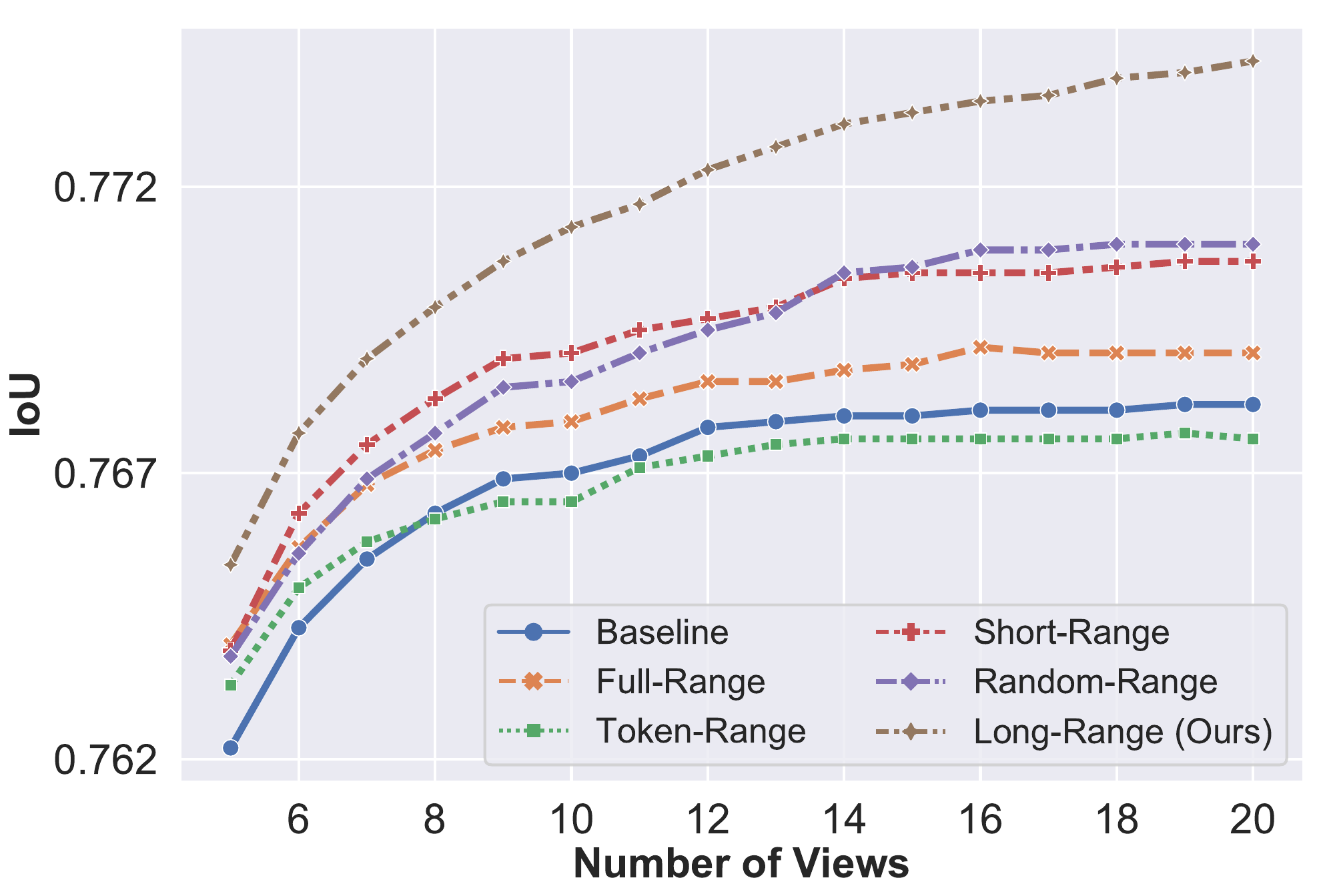}}
	\caption{Comparison of the performance between different grouping strategies when facing various numbers of input views. To control the variables, all experiments employ the same decoder as LRGT and uniformly do not use IFS in the encoder. These experiments are based on ShapeNet and the evaluation metric is IoU.}
	\label{group_type}
\end{figure}

Figure~\ref{group_type} shows the reconstruction quantitative results for the methods using different grouping strategies. The baseline method indicates the model that extracts the feature from views independently while without any inter-view communications. The other control methods are obtained by replacing LGA in our proposed LRGT with the specified attention layer. Note that, the full-range attention here is not used on all transformer blocks, so its performance is in a normal state. Actually, we find in extra experiments that the performance of the model which uses full-range attention in every transformer block is terrible and far worse than the models shown in this figure. It is consistent with our previous analysis. In addition, random-range grouping attention is a special experiment that groups the tokens randomly and evenly.

The model with full-range attention exhibits superior overall performance relative to the baseline. It verifies that establishing inter-view associations is indeed conducive to feature representation. However, the performance of the method using full-range attention tends to be saturated or even degraded as the number of input views gradually increases (especially on 16-20 views). Too many tokens from views bring great inference difficulty to the attention operation. The method with token-range grouping attention is even worse than the baseline because the tokens that establish the association between views are weak in representing the views they located.

The remaining three grouping attention methods have achieved relatively good performance, and these attention layers can save about $98\%$ of the computational complexity compared to the full-range attention layer when facing 20-view input. The method using short-range grouping attention focuses on the local information change between different views. It is difficult to construct non-local correlations between intra-view tokens. Therefore, the multi-view input without temporal coherence in our task is not suitably handled by this method. The random-range grouping attention has the potential to represent the macro feature of each view in the groups, however, it is not stable enough due to random. Only long-range grouping attention can provide reliable macro information in the group of attention operation relatively stably. As a result, the method using long-range grouping attention consistently exhibits superior performance compared to other grouping types whatever the number of view inputs. Even for a large number of view inputs (e.g., 18 to 20 views), it still demonstrates a more substantial performance improvement than other grouping types. Therefore, we consider that the long-range grouping attention is more appropriate for our task.

\begin{table}[]
	\centering
	\renewcommand\arraystretch{1.2}
	\scalebox{0.7}{
    \begin{tabular}{cc|cccccc}
	    \hline
	    \multicolumn{1}{c}{\textbf{LGA}}&
		\multicolumn{1}{c|}{\textbf{IFS}}&
		\multicolumn{1}{c}{\textbf{3 view}} & \multicolumn{1}{c}{\textbf{5 views}} & \multicolumn{1}{c}{\textbf{8 views}} & \multicolumn{1}{c}{\textbf{12 views}} & \multicolumn{1}{c}{\textbf{16 views}} & \multicolumn{1}{c}{\textbf{20 views}} \\ \hline
		\XSolidBrush &\XSolidBrush & 0.7539 & 0.7622 & 0.7663 & 0.7678 & 0.7681 & 0.7682 \\
		\Checkmark &\XSolidBrush & 0.7552 & 0.7654 & 0.7699 & 0.7723 & 0.7735 & 0.7742 \\
		\Checkmark &\Checkmark & \textbf{0.7590} & \textbf{0.7692} & \textbf{0.7744} & \textbf{0.7766} & \textbf{0.7781} & \textbf{0.7786} \\ \hline
	\end{tabular}}
	\caption{Ablation experiments on the effect of long-range grouping attention (LGA) and inter-view feature signatures (IFS). The purpose of IFS is to assist LGA, thereby LGA is preserved in ablation experiments about IFS. The experiments are based on ShapeNet and the evaluation metrics is IoU.}
\label{encoder_ablation}
\end{table}

\subsection{Ablation Study}
\label{experiments::ablation}
\subsubsection{Encoder}
The results of the ablation experiment on our encoder are shown in Table~\ref{encoder_ablation}.

\textbf{Effect of LGA.} To validate the effect of LGA, we replace it with the original multi-head self-attention that handles the branches of each view independently. It leads to a notable decline in performance, especially when the number of input views is extensive. Specifically, the performance decrease is more pronounced for facing 16 views ($0.70\%$) or 20 views ($0.78\%$) than facing 5 views ($0.42\%$). It shows that LGA plays an essential role in exploring multi-view relationships. As the number of views increases, the performance improved after using LGA becomes increasingly evident. It demonstrates that LGA is beneficial to establishing communication among views.

\textbf{Effect of IFS.} To validate the effect of IFS, we attempt to remove it from the encoder. As a result, there is a certain decline in performance. Furthermore, the results reflect that the reconstruction performance reduces more significantly as the number of input views enlarges. The performance drops $0.49\%$ when facing 5 views while dropping $0.53\%$ when facing 20 views input. It verifies that the IFS enhances the inter-view differences to assist with inter-view tokens association.

\begin{figure}[t]
	\centering
	\scalebox{1.25}{
	\includegraphics[width=0.77\linewidth]{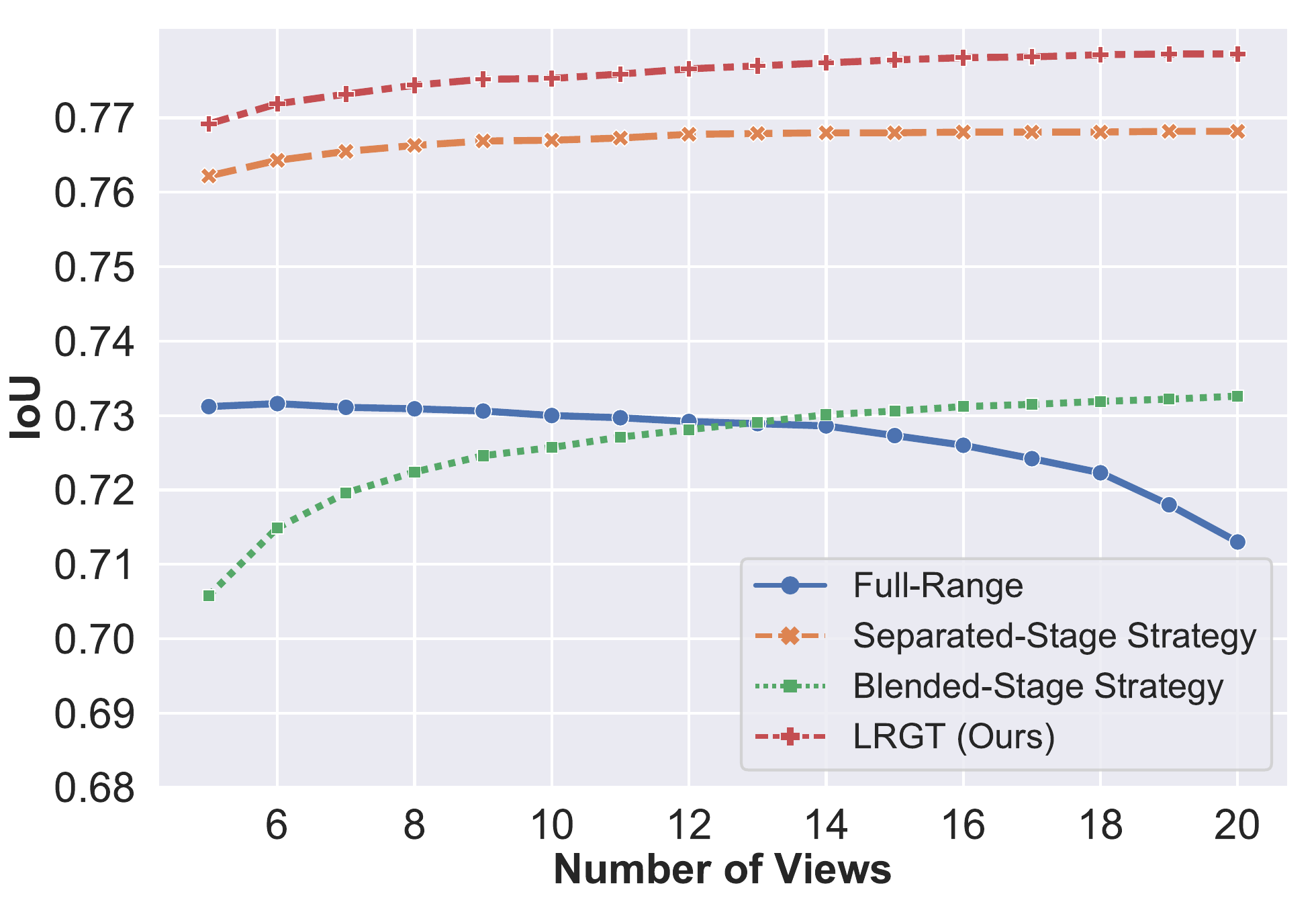}}
	\caption{Comparison of the reconstruction performance between different encoder strategies on the test set of ShapeNet evaluated by IoU. To control the variables, the experiments utilize the same decoder as LRGT.}
	\label{strategy2}
\end{figure}

\textbf{Effect of Encoder Strategies.} As mentioned in Section~\ref{Introduction}, the intuitive approach to introduce the transformer paradigm to extract features from multi-view input is to adopt FRA. However, it is hard for the model to predict the correlation between tokens when facing a heavy amount of input due to the curse of information content. As shown in Figure~\ref{strategy2}, the method with FRA performs rather terribly on multi-view reconstruction. Especially, its performance gradually degrades when increasing the number of inputs over 6. Methods using the separated-stage strategy and blended-stage strategy can alleviate the problems caused by FRA, however, our LRGT can lead to better performance.

\begin{table}[]
	\centering
	\renewcommand\arraystretch{1.3}
	\scalebox{0.7}{
    \begin{tabular}{c|cccccc}
	    \hline
		\multicolumn{1}{c|}{\textbf{\makecell{Skip\\ Connection}}} &
		\multicolumn{1}{c}{\textbf{3 views}} & \multicolumn{1}{c}{\textbf{5 views}} & \multicolumn{1}{c}{\textbf{8 views}} & \multicolumn{1}{c}{\textbf{12 views}} & \multicolumn{1}{c}{\textbf{16 views}} & \multicolumn{1}{c}{\textbf{20 views}}\\ \hline
		\XSolidBrush & 0.7573 & 0.7668 & 0.7717 & 0.7735 & 0.7746 & 0.7752 \\
		\Checkmark & \textbf{0.7590} & \textbf{0.7692} & \textbf{0.7744} & \textbf{0.7766} & \textbf{0.7781} & \textbf{0.7786} \\ \hline
	\end{tabular}}
	\caption{Ablation experiments on the effect of skip connection in HR basic unit of decoder. The experiments are based on ShapeNet dataset and the evaluation metrics is IoU.}
\label{decoder_ablation}
\end{table}

\begin{table}[]
	\centering
	\renewcommand\arraystretch{1.25}
	\scalebox{0.69}{
    \begin{tabular}{c|cccccc}
	    \hline
		\multicolumn{1}{c|}{\textbf{Decoder Type}} & \multicolumn{1}{c}{\textbf{3 views}} & \multicolumn{1}{c}{\textbf{5 views}} & \multicolumn{1}{c}{\textbf{8 views}} & \multicolumn{1}{c}{\textbf{12 views}} & \multicolumn{1}{c}{\textbf{16 views}}  & \multicolumn{1}{c}{\textbf{20 views}}\\\hline
		\textbf{\textbf{EVolT}} & 0.7576 & 0.7677 & 0.7724 & 0.7744 & 0.7759 & 0.7765 \\
		\textbf{\textbf{LegoFormer}} & 0.7513 & 0.7614 & 0.7664 & 0.7693 & 0.7708 & 0.7711 \\
		\textbf{\textbf{3D-RETR}} & 0.7511 & 0.7608 & 0.7650 & 0.7669 & 0.7680 & 0.7681 \\
        \textbf{\textbf{Ours}} & \textbf{0.7590} & \textbf{0.7692} & \textbf{0.7744} & \textbf{0.7766} & \textbf{0.7781} & \textbf{0.7786} \\ \hline
	\end{tabular}}
	\caption{Comparison of the performance between our decoder and others from prior works (EVolT~\cite{wang2021multi}, LegoFormer~\cite{yagubbayli2021legoformer}, 3D-RETR~\cite{shi20213d}).  To control the variables, all experiments employ the same encoder as LRGT. The experiments are based on ShapeNet and the evaluation metrics is IoU.}
\label{decoder_type}
\end{table}

\begin{figure*}
  \centering
	\scalebox{1}{
	\includegraphics[width=1\linewidth]{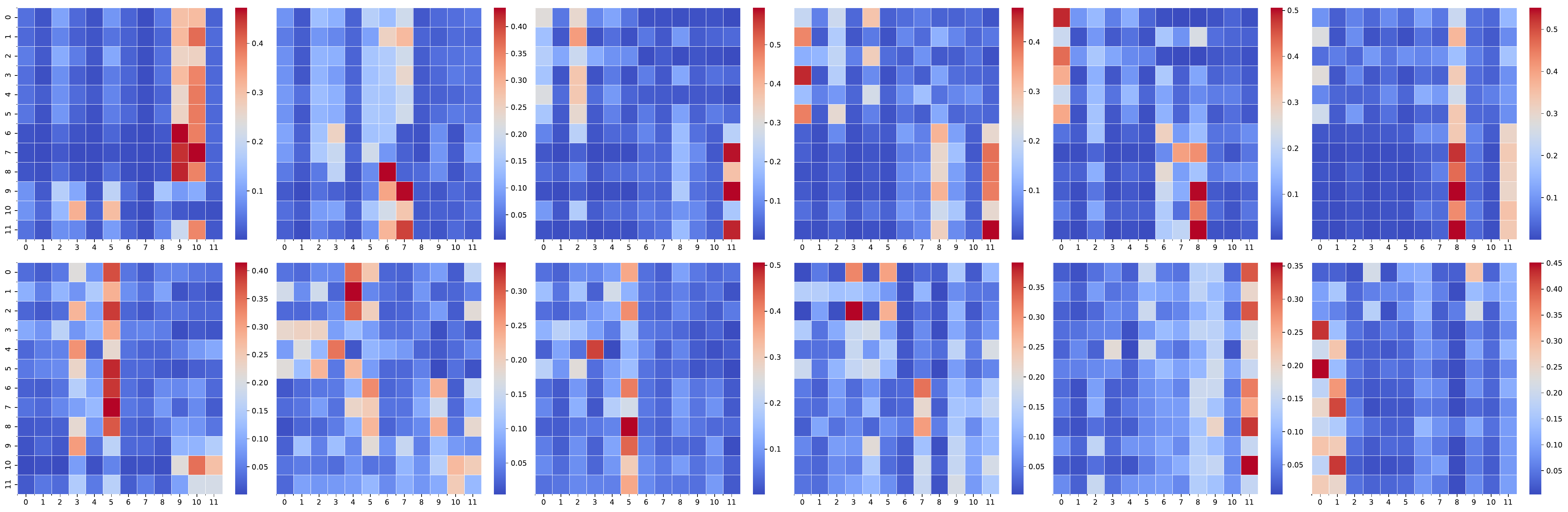}}
    \caption{Visualization of the attention weight maps from different groups in the $1$-st head of the $2$-nd LGA when processing 3-view input. There is a significant difference in the regions concerned by the attention operations between the groups, which ensures a diversity of the overall features. The complete maps are provided in the supplementary material.} 
  \label{heatmap}
\end{figure*}

\subsubsection{Decoder}
First of all, we should validate the effect of skip connection in the HR basic units of our decoder. As shown in Table~\ref{decoder_ablation}, the reconstruction performance drops significantly when the skip connection is discarded. It indicates that the advantages of convolutional layers for local representations make up for the loss caused by combining adjacent voxels in the transformer block with the assistance of this structure.

Besides, we compare the reconstruction performance of our decoder with the decoders proposed in prior advance works \cite{wang2021multi, yagubbayli2021legoformer, shi20213d}. As shown in Table~\ref{decoder_type}, our decoder outperforms the others for multi-view 3D reconstruction undoubtedly. Their methods firstly obtain a voxel feature with the size of $4^3$ with lots of transformer blocks and then upsample it to the target resolution in hasty steps using simple structures such as several fully connected layers or convolution layers. It does not take full advantage of the powerful transformer structure to help the upsampling process which is hard to learn. However, we employ a progressive upsampling architecture which reduces the training difficulty by breaking down the upsampling process. The transformer blocks are distributed to each upsampling stage to achieve better representation.

\subsection{Visualization of Grouping Diversity} 

In Figure~\ref{heatmap}, we visualize several attention weight maps from different groups in an LGA. Each map represents the importance scores between the tokens in a group. In one map, the horizontal axis represents all tokens within this group and the weights refer to the relationship between tokens from the group across views. The part with higher weight means that gains more attention. Due to the grouping mechanism, the corresponding tokens in various groups have proximity positions in the image and similar semantics. However, the weight distribution of the groups with similar semantic arrangements exhibits significant diversity in different maps. It implies that the features get connection following diverse patterns in distinct groups of LGA. Therefore, this structure has a strong representation ability.

\begin{table}[]
\centering
\renewcommand\arraystretch{1.1}
\scalebox{0.75}{
\begin{tabular}{c|c|c|c|c|c}
    \hline
    \multicolumn{6}{c}{\textbf{IoU}} \\ \hline
    \textbf{Pix2Vox++} & \textbf{3D-RETR} & \textbf{GARNet} & \textbf{UMIFormer} & \textbf{LRGT} & \textbf{LRGT$^{\dagger}$}\\ \hline
    0.279 & 0.297 & 0.291 & 0.300 & 0.299 & \textbf{0.304} \\ \hline \hline
    \multicolumn{6}{c}{\textbf{F-score$@1\%$}} \\ \hline
    \textbf{Pix2Vox++} & \textbf{3D-RETR} & \textbf{GARNet} & \textbf{UMIFormer} & \textbf{LRGT} & \textbf{LRGT$^{\dagger}$} \\ \hline
    0.113 & 0.125 & 0.116 & 0.129 & 0.127 & \textbf{0.130} \\ \hline
\end{tabular}}
\caption{Comparison of single-view object reconstruction results on Pix3D using IoU and F-score@$1\%$. $``^{\dagger}"$ means the original transformer blocks replace the LGA transformer blocks in LRGT and other structures are consistent with LRGT.}
\label{pix3d_results}
\end{table}

\subsection{Evaluation on the Pix3D Dataset}
It is crucial to evaluate the ability of the proposed methods to handle real-world data. Therefore, we measure our methods on the Pix3D dataset, which provides single-view reconstruction testing with real-world view images. In detail, following \cite{xie2019pix2vox,xie2020pix2vox++}, we generate the training set using the data from the category of Chair in ShapeNet and synthesize images~\cite{su2015render} with random background from the SUN database~\cite{xiao2010sun}. Each object has 60 synthesized images.

\begin{figure}[ht]
	\centering
	\scalebox{1.25}{
	\includegraphics[width=0.8\linewidth]{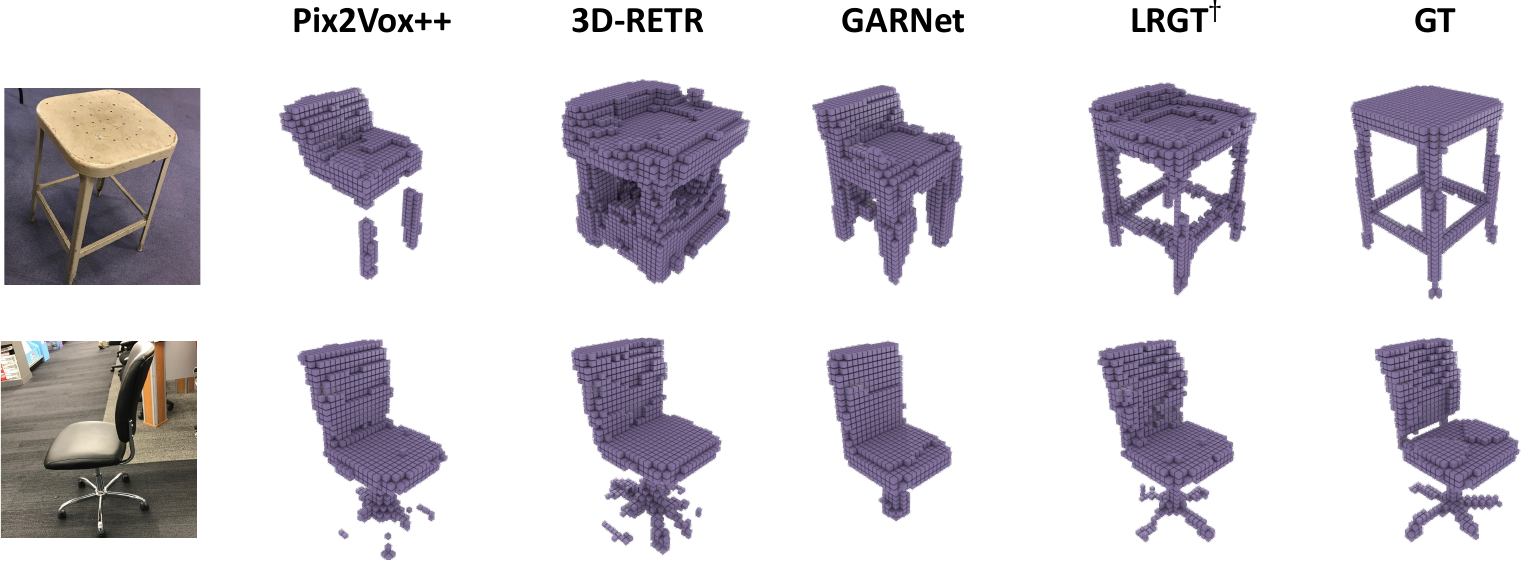}}
	\caption{Single-view reconstruction results on Pix3D.}
	\label{show_results_pix3d}
\end{figure}

We provide two models, LRGT and LRGT$^{\dagger}$. In LRGT$^{\dagger}$, LGA and IFS are removed and replaced by the original self-attention layer while the other structure is the same as in LRGT. Table~\ref{pix3d_results} shows the performance of LRGT, LRGT$^{\dagger}$, and other methods~\cite{xie2020pix2vox++, shi20213d, zhu2023garnet, zhu2023umi}. As LGA and IFS are explicitly designed for multi-view input and Pix3D only provides a single view of each object, the performance of LRGT is slightly lower than LRGT$^{\dagger}$. Nevertheless, the performance of LRGT is still comparable to previous methods. LRGT$^{\dagger}$ outperforms other methods, which shows the excellent performance of our decoder to reconstruct real-world data. The qualitative examples as shown in Figure~\ref{show_results_pix3d} verify that our method outperforms the others in capturing fine details of chairs.

\section{Conclusions and Limitations}

In this paper, we propose a novel transformer-based network for multi-view 3D reconstruction which achieves SOTA accuracy. Its encoder extracts the reliable features from multi-view employing the LGA to establish communication between views and the IFS to enhance the feature differences between various views. Besides, a progressive upsampling decoder is designed for powerful voxel reconstruction performance. In conclusion, our method has corresponding contributions in handling multi-view input and generating 3D objects. Therefore, we expected that these approaches can be introduced to other suitable application scenarios. The major limitation of our proposed method is that the group strategy in LGA method adheres to a fixed pattern. Although it simplifies the problem compared to the FRA method, the relationship between different views still needs to rely on network learning. We expect to continue optimizing this scheme with learnable grouping in future work.

~\\
\setlength{\parindent}{0pt} { \textbf{Acknowledgements} This work was supported by National Key Research and Development Plan under Grant 2021YFE0205700, Science and Technology Development Fund of Macau (0070/2020/AMJ, 0004/2020/A1) and Guangdong Provincial Key R\&D Programme: 2019B010148001.
}
{\small
\bibliographystyle{unsrt}
\bibliography{reference}

\begin{thebibliography}{10}

\bibitem{choy20163d}
Christopher~B Choy, Danfei Xu, JunYoung Gwak, Kevin Chen, and Silvio Savarese.
\newblock 3d-r2n2: A unified approach for single and multi-view 3d object
  reconstruction.
\newblock In {\em European conference on computer vision}, pages 628--644.
  Springer, 2016.

\bibitem{kar2017learning}
Abhishek Kar, Christian H{\"a}ne, and Jitendra Malik.
\newblock Learning a multi-view stereo machine.
\newblock {\em Advances in neural information processing systems}, 30, 2017.

\bibitem{su2015multi}
Hang Su, Subhransu Maji, Evangelos Kalogerakis, and Erik Learned-Miller.
\newblock Multi-view convolutional neural networks for 3d shape recognition.
\newblock In {\em Proceedings of the IEEE international conference on computer
  vision}, pages 945--953, 2015.

\bibitem{wang20173densinet}
Meng Wang, Lingjing Wang, and Yi~Fang.
\newblock 3densinet: A robust neural network architecture towards 3d volumetric
  object prediction from 2d image.
\newblock In {\em Proceedings of the 25th ACM international conference on
  Multimedia}, pages 961--969, 2017.

\bibitem{huang2018deepmvs}
Po-Han Huang, Kevin Matzen, Johannes Kopf, Narendra Ahuja, and Jia-Bin Huang.
\newblock Deepmvs: Learning multi-view stereopsis.
\newblock In {\em Proceedings of the IEEE Conference on Computer Vision and
  Pattern Recognition}, pages 2821--2830, 2018.

\bibitem{paschalidou2018raynet}
Despoina Paschalidou, Osman Ulusoy, Carolin Schmitt, Luc Van~Gool, and Andreas
  Geiger.
\newblock Raynet: Learning volumetric 3d reconstruction with ray potentials.
\newblock In {\em Proceedings of the IEEE Conference on Computer Vision and
  Pattern Recognition}, pages 3897--3906, 2018.

\bibitem{xie2019pix2vox}
Haozhe Xie, Hongxun Yao, Xiaoshuai Sun, Shangchen Zhou, and Shengping Zhang.
\newblock Pix2vox: Context-aware 3d reconstruction from single and multi-view
  images.
\newblock In {\em Proceedings of the IEEE/CVF international conference on
  computer vision}, pages 2690--2698, 2019.

\bibitem{xie2020pix2vox++}
Haozhe Xie, Hongxun Yao, Shengping Zhang, Shangchen Zhou, and Wenxiu Sun.
\newblock Pix2vox++: Multi-scale context-aware 3d object reconstruction from
  single and multiple images.
\newblock {\em International Journal of Computer Vision}, 128(12):2919--2935,
  2020.

\bibitem{yang2020robust}
Bo~Yang, Sen Wang, Andrew Markham, and Niki Trigoni.
\newblock Robust attentional aggregation of deep feature sets for multi-view 3d
  reconstruction.
\newblock {\em International Journal of Computer Vision}, 128(1):53--73, 2020.

\bibitem{zhu2023garnet}
Zhenwei Zhu, Liying Yang, Xuxin Lin, Lin Yang, and Yanyan Liang.
\newblock Garnet: Global-aware multi-view 3d reconstruction network and the
  cost-performance tradeoff.
\newblock {\em Pattern Recognition}, 142:109674, 2023.

\bibitem{wang2021multi}
Dan Wang, Xinrui Cui, Xun Chen, Zhengxia Zou, Tianyang Shi, Septimiu Salcudean,
  Z~Jane Wang, and Rabab Ward.
\newblock Multi-view 3d reconstruction with transformers.
\newblock In {\em Proceedings of the IEEE/CVF International Conference on
  Computer Vision}, pages 5722--5731, 2021.

\bibitem{yagubbayli2021legoformer}
Farid Yagubbayli, Alessio Tonioni, and Federico Tombari.
\newblock Legoformer: Transformers for block-by-block multi-view 3d
  reconstruction.
\newblock {\em arXiv preprint arXiv:2106.12102}, 2021.

\bibitem{shi20213d}
Zai Shi, Zhao Meng, Yiran Xing, Yunpu Ma, and Roger Wattenhofer.
\newblock 3d-retr: End-to-end single and multi-view 3d reconstruction with
  transformers.
\newblock In {\em British Machine Vision Conference (BMVC)}, 2021.

\bibitem{zhu2023umi}
Zhenwei Zhu, Liying Yang, Ning Li, Chaohao Jiang, and Yanyan Liang.
\newblock Umiformer: Mining the correlations between similar tokens for
  multi-view 3d reconstruction.
\newblock {\em arXiv preprint arXiv:2302.13987}, 2023.

\bibitem{dosovitskiy2021image}
Alexey Dosovitskiy, Lucas Beyer, Alexander Kolesnikov, Dirk Weissenborn,
  Xiaohua Zhai, Thomas Unterthiner, Mostafa Dehghani, Matthias Minderer, Georg
  Heigold, Sylvain Gelly, et~al.
\newblock An image is worth 16x16 words: Transformers for image recognition at
  scale.
\newblock In {\em International Conference on Learning Representations}, 2021.

\bibitem{zeng2020learning}
Yanhong Zeng, Jianlong Fu, and Hongyang Chao.
\newblock Learning joint spatial-temporal transformations for video inpainting.
\newblock In {\em European Conference on Computer Vision}, pages 528--543.
  Springer, 2020.

\bibitem{liu2021decoupled}
Rui Liu, Hanming Deng, Yangyi Huang, Xiaoyu Shi, Lewei Lu, Wenxiu Sun, Xiaogang
  Wang, Jifeng Dai, and Hongsheng Li.
\newblock Decoupled spatial-temporal transformer for video inpainting.
\newblock {\em arXiv preprint arXiv:2104.06637}, 2021.

\bibitem{chang2015shapenet}
Angel~X Chang, Thomas Funkhouser, Leonidas Guibas, Pat Hanrahan, Qixing Huang,
  Zimo Li, Silvio Savarese, Manolis Savva, Shuran Song, Hao Su, et~al.
\newblock Shapenet: An information-rich 3d model repository.
\newblock {\em arXiv preprint arXiv:1512.03012}, 2015.

\bibitem{sun2018pix3d}
Xingyuan Sun, Jiajun Wu, Xiuming Zhang, Zhoutong Zhang, Chengkai Zhang, Tianfan
  Xue, Joshua~B Tenenbaum, and William~T Freeman.
\newblock Pix3d: Dataset and methods for single-image 3d shape modeling.
\newblock In {\em Proceedings of the IEEE conference on computer vision and
  pattern recognition}, pages 2974--2983, 2018.

\bibitem{ozyecsil2017survey}
Onur {\"O}zye{\c{s}}il, Vladislav Voroninski, Ronen Basri, and Amit Singer.
\newblock A survey of structure from motion*.
\newblock {\em Acta Numerica}, 26:305--364, 2017.

\bibitem{fuentes2015visual}
Jorge Fuentes-Pacheco, Jos{\'e} Ruiz-Ascencio, and Juan~Manuel
  Rend{\'o}n-Mancha.
\newblock Visual simultaneous localization and mapping: a survey.
\newblock {\em Artificial intelligence review}, 43(1):55--81, 2015.

\bibitem{lin2021barf}
Chen-Hsuan Lin, Wei-Chiu Ma, Antonio Torralba, and Simon Lucey.
\newblock Barf: Bundle-adjusting neural radiance fields.
\newblock In {\em Proceedings of the IEEE/CVF International Conference on
  Computer Vision}, pages 5741--5751, 2021.

\bibitem{zhang2021ners}
Jason Zhang, Gengshan Yang, Shubham Tulsiani, and Deva Ramanan.
\newblock Ners: Neural reflectance surfaces for sparse-view 3d reconstruction
  in the wild.
\newblock {\em Advances in Neural Information Processing Systems},
  34:29835--29847, 2021.

\bibitem{yang2022fvor}
Zhenpei Yang, Zhile Ren, Miguel~Angel Bautista, Zaiwei Zhang, Qi~Shan, and
  Qixing Huang.
\newblock Fvor: Robust joint shape and pose optimization for few-view object
  reconstruction.
\newblock In {\em Proceedings of the IEEE/CVF Conference on Computer Vision and
  Pattern Recognition}, pages 2497--2507, 2022.

\bibitem{vaswani2017attention}
Ashish Vaswani, Noam Shazeer, Niki Parmar, Jakob Uszkoreit, Llion Jones,
  Aidan~N Gomez, {\L}ukasz Kaiser, and Illia Polosukhin.
\newblock Attention is all you need.
\newblock {\em Advances in neural information processing systems}, 30, 2017.

\bibitem{Fwang2020axial}
Huiyu Wang, Yukun Zhu, Bradley Green, Hartwig Adam, Alan Yuille, and
  Liang-Chieh Chen.
\newblock Axial-deeplab: Stand-alone axial-attention for panoptic segmentation.
\newblock In {\em Computer Vision--ECCV 2020: 16th European Conference,
  Glasgow, UK, August 23--28, 2020, Proceedings, Part IV}, pages 108--126.
  Springer, 2020.

\bibitem{huang2019interlaced}
Lang Huang, Yuhui Yuan, Jianyuan Guo, Chao Zhang, Xilin Chen, and Jingdong
  Wang.
\newblock Interlaced sparse self-attention for semantic segmentation.
\newblock {\em arXiv preprint arXiv:1907.12273}, 2019.

\bibitem{dong2022cswin}
Xiaoyi Dong, Jianmin Bao, Dongdong Chen, Weiming Zhang, Nenghai Yu, Lu~Yuan,
  Dong Chen, and Baining Guo.
\newblock Cswin transformer: A general vision transformer backbone with
  cross-shaped windows.
\newblock In {\em Proceedings of the IEEE/CVF Conference on Computer Vision and
  Pattern Recognition}, pages 12124--12134, 2022.

\bibitem{park2022how}
Namuk Park and Songkuk Kim.
\newblock How do vision transformers work?
\newblock In {\em International Conference on Learning Representations}, 2022.

\bibitem{milletari2016v}
Fausto Milletari, Nassir Navab, and Seyed-Ahmad Ahmadi.
\newblock V-net: Fully convolutional neural networks for volumetric medical
  image segmentation.
\newblock In {\em 2016 fourth international conference on 3D vision (3DV)},
  pages 565--571. IEEE, 2016.

\bibitem{tatarchenko2019single}
Maxim Tatarchenko, Stephan~R Richter, Ren{\'e} Ranftl, Zhuwen Li, Vladlen
  Koltun, and Thomas Brox.
\newblock What do single-view 3d reconstruction networks learn?
\newblock In {\em Proceedings of the IEEE/CVF conference on computer vision and
  pattern recognition}, pages 3405--3414, 2019.

\bibitem{lorensen1987marching}
William~E Lorensen and Harvey~E Cline.
\newblock Marching cubes: A high resolution 3d surface construction algorithm.
\newblock {\em ACM siggraph computer graphics}, 21(4):163--169, 1987.

\bibitem{touvron2021training}
Hugo Touvron, Matthieu Cord, Matthijs Douze, Francisco Massa, Alexandre
  Sablayrolles, and Herv{\'e} J{\'e}gou.
\newblock Training data-efficient image transformers \& distillation through
  attention.
\newblock In {\em International conference on machine learning}, pages
  10347--10357. PMLR, 2021.

\bibitem{loshchilov2018decoupled}
Ilya Loshchilov and Frank Hutter.
\newblock Decoupled weight decay regularization.
\newblock In {\em International Conference on Learning Representations}, 2019.

\bibitem{su2015render}
Hao Su, Charles~R Qi, Yangyan Li, and Leonidas~J Guibas.
\newblock Render for cnn: Viewpoint estimation in images using cnns trained
  with rendered 3d model views.
\newblock In {\em Proceedings of the IEEE international conference on computer
  vision}, pages 2686--2694, 2015.

\bibitem{xiao2010sun}
Jianxiong Xiao, James Hays, Krista~A Ehinger, Aude Oliva, and Antonio Torralba.
\newblock Sun database: Large-scale scene recognition from abbey to zoo.
\newblock In {\em 2010 IEEE computer society conference on computer vision and
  pattern recognition}, pages 3485--3492. IEEE, 2010.

\end{thebibliography}
}

\end{document}